\documentclass{article}

\usepackage{microtype}
\usepackage{graphicx}
\usepackage{booktabs} 

\usepackage{amsmath,amsfonts,bm}

\def\eqref#1{equation~\ref{#1}}

\def\1{\bm{1}}

\DeclareMathAlphabet{\mathsfit}{\encodingdefault}{\sfdefault}{m}{sl}
\SetMathAlphabet{\mathsfit}{bold}{\encodingdefault}{\sfdefault}{bx}{n}

\newcommand{\E}{\mathbb{E}}

\usepackage{hyperref}
\usepackage{url}
\usepackage{enumitem}
\usepackage{multirow}
\usepackage{wrapfig}
\usepackage{comment}
\usepackage{setspace}
\usepackage{balance}
\usepackage{subcaption}
\usepackage{longtable}
\usepackage{multicol}
\usepackage{hyperref}

\usepackage[accepted]{icml2023}

\usepackage{amsmath}
\usepackage{amssymb}
\usepackage{mathtools}
\usepackage{amsthm}
\usepackage{algorithm}
\usepackage{algorithmic}

\usepackage[capitalize,noabbrev]{cleveref}

\theoremstyle{plain}

\theoremstyle{definition}

\theoremstyle{remark}

\usepackage[textsize=tiny]{todonotes}

\icmltitlerunning{Probabilistic Categorical Adversarial Attack and Adversarial Training}

\begin{document}

\twocolumn[
\icmltitle{Probabilistic Categorical Adversarial Attack and  Adversarial Training}



\icmlsetsymbol{equal}{*}

\begin{icmlauthorlist}
\icmlauthor{Han Xu}{msu}
\icmlauthor{Pengfei He}{msu}
\icmlauthor{Jie Ren}{msu}
\icmlauthor{Yuxuan Wan}{msu}
\icmlauthor{Zitao Liu}{tal}
\icmlauthor{Hui Liu}{msu}
\icmlauthor{Jiliang Tang}{msu}
\end{icmlauthorlist}

\icmlaffiliation{msu}{Department of Computer Science and Engineering, Michigan State University, East Lansing, MI, USA}
\icmlaffiliation{tal}{Guangdong Institute of Smart Education, Jinan University, Guangzhou, China}

\icmlcorrespondingauthor{Zitao Liu}{liuzitao@jnu.edu.cn}

\icmlkeywords{Machine Learning, ICML}

\vskip 0.3in
]



\printAffiliationsAndNotice{} 

\newcommand{\method}{\textsc{PCAA~}}
\newcommand{\defense}{\textsc{PAdvT~}}

\newcommand{\fix}{\marginpar{FIX}}
\newcommand{\new}{\marginpar{NEW}}
\newcommand{\pf}[1]{\textcolor{orange}{#1}}
\newcommand{\han}[1]{\textcolor{red}{#1}}
\newcommand{\yx}[1]{\textcolor{blue}{#1}}
\newcommand{\jt}[1]{\textcolor{blue}{#1}}

\begin{abstract} 
The studies on adversarial attacks and defenses have greatly improved the robustness of  Deep Neural Networks (DNNs). Most advanced approaches have been overwhelmingly designed for continuous data such as images. However, these achievements are still hard to be generalized to categorical data.
To bridge this gap, we propose a novel framework, \textit{Probabilistic Categorical Adversarial Attack (or PCAA)}. It transfers the discrete optimization problem of finding categorical adversarial examples to a continuous problem that can be solved via gradient-based methods. We analyze the optimality  (attack success rate) and time complexity of \method to demonstrate its significant advantage over current search-based attacks. More importantly, through extensive empirical studies, we demonstrate that the well-established defenses for continuous data, such as adversarial training and TRADES, can be easily accommodated to defend DNNs for categorical data.
\end{abstract}

\section{Introduction}

Adversarial examples~\cite{DBLP:journals/corr/GoodfellowSS14} have raised great concerns for the applications of Deep Neural Networks (DNNs) in many security-critical domains ~\cite{DBLP:journals/adhoc/CuiLSZ19, DBLP:conf/acsac/StringhiniKV10, DBLP:journals/nca/CaoT01}. Recent years have witnessed an increasing number of adversarial attack and defense methods~\cite{DBLP:journals/corr/GoodfellowSS14, DBLP:conf/iclr/MadryMSTV18,ilyas2019adversarial}. These studies have not only greatly deepened our understanding on the vulnerabilities of DNNs but also tremendously advanced the robustness of DNNs. Until now, the majority of existing accomplishments have been achieved in continuous data such as images, where gradient-based approaches can be leveraged. 

However, there are also many machine learning tasks where the input data is categorical. For example, data in ML-based intrusion detection systems \cite{DBLP:journals/cybersec/KhraisatGVK19} contains records of the type of system operations; in financial transaction systems, data includes categorical features such as the region and card information of transactions; and in NLP tasks, the words in a sentence can only be chosen from a given vocabulary, which is categorical. To generate categorical adversarial examples, there are recent search-based approaches such as~\cite{DBLP:journals/jmlr/YangCHWJ20, DBLP:conf/mlsys/LeiWCDDW19, DBLP:conf/iclr/BaoHZS022}. For example, the method~\cite{DBLP:journals/jmlr/YangCHWJ20} first finds top-$K$ features of a given sample that have the maximal influence on the model output, and then a greedy search is applied to obtain the optimal perturbation in these $K$ features. However, these search-based attack methods usually suffer from a poor trade-off between efficiency and optimality (attack success rate) to find strong adversarial examples. 
Moreover, if these attack methods are applied in defenses like adversarial training~\cite{DBLP:conf/iclr/MadryMSTV18}, they can only search for adversarial examples for each training sample at each time, instead of efficiently producing adversarial examples in batches. In a nutshell, these drawbacks of existing categorical attack methods dramatically prohibit the possibility of applying recent advances of attack and defense established in continuous data to categorical data.

Therefore, a natural question is \textit{can we generalize the well-studied methods of continuous data to categorical data?} We face tremendous challenges to answer this question. First, the input data space are categorical, thus the gradient methods of adversarial attacks and defenses~\cite{DBLP:journals/corr/GoodfellowSS14, DBLP:conf/iclr/MadryMSTV18} for continuous data are not directly applicable. Second, most attacks for categorical data desire to constrain the number of perturbed features~\cite{DBLP:journals/jmlr/YangCHWJ20, DBLP:conf/iclr/BaoHZS022}, which is different from the commonly considered $l_2,l_{\infty}$ norm constraints in continuous data. To address these challenges, we propose a novel framework: \textbf{P}robabilistic \textbf{C}ategorical \textbf{A}dversarial \textbf{A}ttack (PCAA). Overall, it transforms the discrete optimization problem of finding categorical adversarial examples into a continuous problem by estimating the probabilistic distribution of categorical adversarial examples. In detail, given a clean data sample, we assume that (each feature of) its adversarial example follows a categorical distribution, and satisfies: \textbf{(1)} the samples following this distribution have a high expected loss value and \textbf{(2)} the samples only have a few features which are different from the original clean sample (with high probability). Based on this property, we are able to leverage existing gradient-based algorithms from continuous data such as ~\cite{DBLP:journals/corr/GoodfellowSS14, DBLP:conf/iclr/MadryMSTV18} to figure out the adversarial examples for categorical data. In such a way, we can successfully obtain the categorical adversarial examples by optimizing the adversarial distribution and taking samples from this distribution.
Meanwhile, our attack can also be easily incorporated with existing powerful defenses for continuous data such as adversarial training \cite{DBLP:conf/iclr/MadryMSTV18} and TRADES~\cite{tu2019theoretical}.
Empirically, we verify that our attack can achieve a better optimality vs.~efficiency trade-off to find strong adversarial examples, and demonstrate the advantages of our defense over other categorical defenses.
To summarize, the major contributions of this paper are:
\begin{itemize}
  \setlength\itemsep{0em}
    \item We propose an efficient and effective framework \method to bridge the gap between categorical data and continuous data, which allows us to generate adversarial examples for categorical data by leveraging methods from continuous data.
    \item Equipped with \method, existing defenses in continuous data, such as adversarial training and TRADES, can be easily adapted to categorical data. This contribution enables us to generalize new advances in defenses from continuous data to categorical data.
    \item We empirically validate the great benefit of \method from perspectives of both attack and defense.
\end{itemize}
\section{Related Work}
In this section, we provide a brief review of existing methodologies of adversarial attacks and  defenses for continuous and categorical data, which highlights the gap between the major methodologies in these two types of data.

\subsection{Attacks and Defenses on Continuous Data}

The adversarial attacks and  defenses in the image domain have been extensively studied \citep{DBLP:conf/iclr/MadryMSTV18, ilyas2019adversarial, DBLP:journals/ijautcomp/XuMLDLTJ20}. Most frequently studied attack methods such as FGSM~\citep{DBLP:journals/corr/GoodfellowSS14}, PGD~\citep{DBLP:conf/iclr/MadryMSTV18} and C\&W Attack~\cite{carlini2017towards} can only be conducted in the continuous domain, since they require calculating the gradients of model outputs on the input data. As countermeasures to resist adversarial examples in the image domain, most defense strategies are also based on the assumption that the input data space is continuous. For example, adversarial training methods~\cite{DBLP:conf/iclr/MadryMSTV18, DBLP:journals/corr/abs-1901-08573} train the DNNs on the adversarial examples generated by PGD. A SOTA certified defense method Randomized Smoothing~\cite{cohen2019certified}, calculates the certifiable bounds by adding Gaussian noise to the input samples. However, these methods are hard to be directly applicable to  categorical data as the input data space is discrete.

\subsection{Attacks and Defenses on Categorical Data} 

To generate adversarial examples for categorical data, most existing methods apply a search-based approach. For example, the works \citep{DBLP:journals/jmlr/YangCHWJ20, DBLP:conf/iclr/BaoHZS022, DBLP:conf/mlsys/LeiWCDDW19} first find top-$K$ features of a given sample that have the maximal influence on the model output, and then a greedy search or brutal search is applied to obtain the optimal combination of perturbation in these $K$ features. In NLP tasks, there are also many attack methods proposed to find adversarial ``sentences'' or ``articles'', which follow a similar approach as general search-based categorical attacks. For example, \citet{ebrahimi2017hotflip} proposes to search important characters in the text based on the gradient, and then apply greedy search to find the optimal character flipping. 
\citet{samanta2017towards} generate the adversarial embedding in the word embedding space, then search for the closest meaningful adversarial example that is legitimate in the text domain. These attack methods are very different from those in continuous domain.

To defend against categorical attacks, most defenses have been proposed for NLP tasks and exclusively rely on the property of word embedding: similar words have a close distance in the embedding
space. \citet{miyato2016adversarial, barham2019interpretable} conduct adversarial training on embedding space, where the adversarial examples are within $l_2$-ball around the embedding of clean samples. 
\citet{DBLP:conf/iclr/DongLJ021} also proposes an adversarial training method - ASCC, which conducts adversarial training in the space which is composed of convex hulls of adversarial word vectors. However, these methods can hardly be applied to defend DNNs for general categorical data beyond NLP tasks. Different from existing defense methods in NLP tasks, in this paper, our proposed defense does not rely on the property of word embedding and achieves a similar defense performance to these NLP defenses. Moreover, our defense can be applied in general categorical ML tasks beyond NLP.
\section{Probabilistic~Categorical~Adversarial~Attack}\label{sec:PCAA}

In this section, we first introduce the necessary notations and definitions of our studied problem. Then, we provide the details of our proposed attack framework PCAA. 

\subsection{Problem Setup}

In this work, we consider a classifier $f$ that predicts labels $y\in \mathcal{Y}$ based on categorical inputs $x\in \mathcal{X}$. Each input sample $x$ contains $n$ categorical features, and each feature $x_i$ can be perturbed to take a value from $d$ allowed categories. Namely, we define the space of all allowed perturbed samples of $x$ to be $\mathcal{S}(x)$. 
Meanwhile, to keep the perturbation ``un-noticeable'', we follow the setups of existing works~\cite{DBLP:journals/jmlr/YangCHWJ20, DBLP:conf/iclr/BaoHZS022, DBLP:conf/kdd/WangHBSML020}, to limit the number of perturbed features, which is the $l_0$ distance of clean sample $x$ and the adversarial example $x'$. It is because constraining $l_0$ distance is most intuitive and has a broad interest in categorical data, e.g. only a few nucleotides are mutated in genetics data. Therefore, we formally define the objective of our considered attack to be: given the budget size $\epsilon$, we aim to find an adversarial example $x'$, which maximizes the model's loss value, while has a small $l_0$ distance to $x$:
\begin{align}\label{deter form} 
\max_{x' \in \mathcal{S}(x)} \mathcal{L}(f(x'),y) 
~~\text{s.t.}~~ \|x'-x\|_0 \leq \epsilon.
\end{align}
Notably, the objective in Eq.(\ref{deter form}) is general and it can be applied to find adversarial examples in various applications by accommodating the space $\mathcal{S}(x)$. 
For example, in NLP tasks such as sentiment analysis, we can define the space $\mathcal{S}(x)$ to be the set of sentences where some words in $x$ are changed to their synonyms. In this way, we can keep the semantic meaning of $x$ during attacking. More details about how to get $\mathcal{S}(x)$ in NLP tasks are given in Section \ref{sec:exp_defense}.



\subsection{The Objective of PCAA}

To solve the problem in Eq.(\ref{deter form}), there are existing search-based methods~\citep{DBLP:conf/mlsys/LeiWCDDW19, DBLP:journals/jmlr/YangCHWJ20} to search the adversarial examples, via either a greedy search method or brutal search method. However, both of these two search methods can suffer from poor optimality vs.~efficiency trade-off during attacking. For example, if one conducts a brutal search~\cite{DBLP:conf/mlsys/LeiWCDDW19} to traverse the whole discrete space,  it must cause an extremely high computational cost. Meanwhile, greedy search narrows down the search space so that it usually finds weak adversarial examples (which cannot maximize the loss in Eq.(\ref{deter form})). Moreover, the poor optimality vs.~efficiency trade-off will make these attacks impossible to be incorporated to the most studied defense methods (in the continuous domain), such as adversarial training. 

\begin{figure}
    \centering
    \includegraphics[width = 0.5\textwidth]{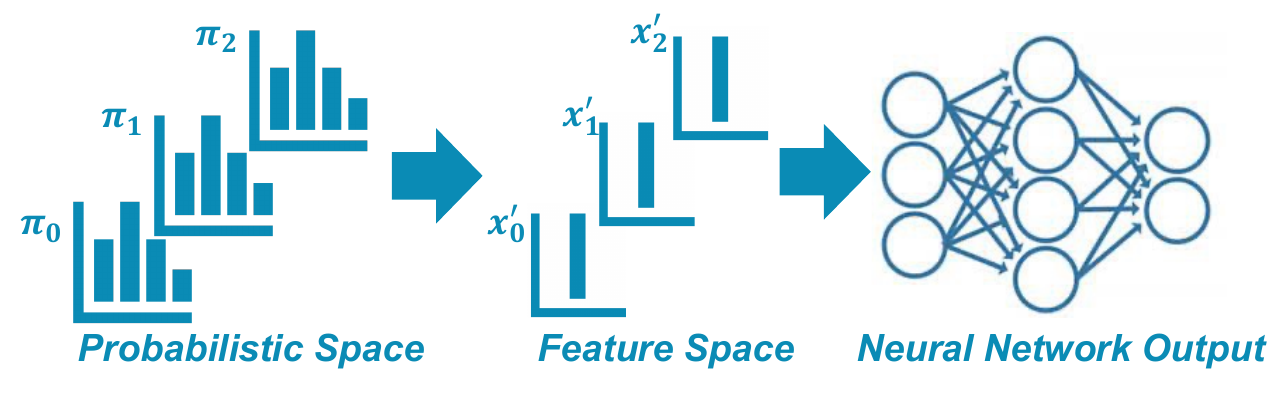}
    \vspace{-0.5cm}
    \caption{An illustration of \method when $n = 3$ and $d = 4$. Each feature $x'_i$ of the adversarial example $x'$ is sampled from probabilistic distribution $\pi_i$ before feed into the model.}
    \label{illustrate}
\end{figure}

To address these problems, we are motivated to leverage gradient-based methods to conduct adversarial attacks in the categorical domain. In general, we first define a \textit{\textbf{continuous probabilistic space}} where the adversarial examples are sampled from, and we devise a new objective in Eq.(\ref{prob cons}) to approximately solve Eq.(\ref{deter form}).
In specific, following the illustration in Figure~\ref{illustrate}, we assume that each feature of (adversarial) categorical data $x'_i$ follows a categorical distribution: $\textit{Categorical}(\pi_i)$, where $\pi_i \in \Pi_i = (0,1)^d$. Each element $\pi_{i,j}$ represents the probability that the feature $i$ takes the category value $j$. 
In the remaining of the paper, we will use $\pi_i$ to denote the categorical distribution $\textit{Categorical}(\pi_i)$ without the loss of generality. 
Then, the input sample $x$'s distribution is the joint distribution of all $\pi_i$, which is denoted as $\pi = [\pi_0; \pi_1; ...;\pi_n] \in \Pi \subset \mathbb{R}^{n\times d}$.

Then, we define a new continuous optimization problem to find a probability distribution $\pi$ in the space of $\Pi$:
\begin{equation}\label{prob cons}
    \max_{\pi\in\Pi} \E_{x'\sim \pi}\mathcal{L}(f(x'),y)~~ \text{s.t.}~~\Pr_{x'\sim\pi}(\|x'-x\|_0\ge \epsilon)\le \delta
\end{equation}
where $\epsilon$ denotes the perturbation budget size and $\delta$ is the tail probability constraint. By solving the problem in Eq.(\ref{prob cons}), we aim to find a distribution with parameter $\pi$ such that: \textbf{(1) \textit{on average}}, the generated samples $x'$ from distribution $\pi$ have a high loss value; and \textbf{(2) \textit{with low probability}}, the sampled $x'$ has a $l_0$ distance to the clean sample $x$ larger than $\epsilon$. Thus, the generated samples $x'$ are likely to mislead the model prediction while preserving most features of $x$. As shown in Figure~\ref{illustrate}, the probabilistic distribution $\pi$ to Eq.(\ref{prob cons}) is first used to sample
adversarial examples $x'$, and then, the model makes predictions on the $x'$.

It worth mentioning that, in our attack, each feature $x'_i$ of the adversarial example $x'$ is sampled  independently. However, it does not mean that the features themselves in the data distribution are independent to each other.  Therefore, our framework is general and applicable to various data types and model architectures, including sequential data such as sentences in NLP areas, or DNA sequences.

\subsection{An Efficient Algorithm of PCAA}
Solving the problem in Eq.(\ref{prob cons}) is not trivial because the probability and the $l_0$ term are not differentiable. Thus, we provide a feasible algorithm to solve Eq.(\ref{prob cons}), by substituting the constraint in Eq.(\ref{prob cons}) to a differentiable term. In detail, we substitute the $l_0$ distance between $x'$ and $x$ by calculating the sum of Cross Entropy Loss between $\pi_i$ and $x_i$, which is $\mathcal{L}_{CE}(\pi_i, x_i)$, for all features $i\in |x|$. It is because  $\mathcal{L}_{CE}(\pi_i, x_i)$ measures the probability that the categorical variables $x'_i$ following the distribution $\pi_i$ is different from $x_i$. Thus, we use the sum of Cross Entropy $\sum_{i\in|n|} \mathcal{L}_{CE}(\pi_i, x_i)$ to approximate the total number of changed features in $x'$, which is the $l_0$ difference $\|x'-x\|_0$. In our algorithm, we penalize the searched $\pi$ when the term $\sum_{i\in|n|} \mathcal{L}_{CE}(\pi_i, x_i)$ exceeds a positive value $\zeta$ as: 
\begin{equation*}\label{eq:step1}
\max_{\pi\in\Pi} \E_{x'\sim \pi}\mathcal{L}(f(x'),y)~~ \text{s.t.}~~
\sum_{i\in |n|} \mathcal{L}_{CE}(x_i, \pi_i) -\zeta \leq 0
\end{equation*}
As a result, we equivalently limit the probability that the generated samples $x'$ have the number of perturbed features larger than $\epsilon$.
Moreover, since the Cross-Entropy Loss is differentiable in terms of $\pi$, we further transform the problem to its Lagrangian form as:
\begin{equation}\label{final optimization}
    \max_{\pi \in \Pi} \Biggl(\E_{x'\sim \pi}\mathcal{L}(f(x'),y)\Biggr) -\lambda \left[ \sum_{i\in |n|}\mathcal{L}_{CE}(x_i, \pi_i)-\zeta\right]^{+}
\end{equation}
where $\lambda$ is the penalty coefficient, and $[\cdot]^{+}$ is $\max(\cdot,0)$. Next, we will show how to solve the maximization problem above by applying gradient methods.

\textbf{Back propagation through Gumbel-Softmax}. Note that the gradient of the expected loss function with respect to $\pi$ cannot be directly calculated in Eq.(\ref{final optimization}), so we apply the Gumbel-Softmax estimator~\cite{DBLP:conf/iclr/JangGP17}. In practice, we consider an unnormalized categorical distribution $\pi_i\in (0,C]^d$, where $C>0$ is a large constant so that the searching space is sufficiently large. The distribution generates sample vectors $x'_i$ as follows: 
\begin{equation}\label{gumbel}
    x'_{ij}=\frac{\exp((\log \pi_{ij}+g_j)/\tau)}{\sum_{j=1}^d\exp((\log \pi_{ij}+g_j)/\tau)},~~\text{for}~j=1,...,d
\end{equation}
where $g_j$ denotes i.i.d samples from the Gumbel$(0,1)$ distribution, and $\tau$ is the softmax temperature. This re-parameterization process facilitates us to calculate the gradient of the expected loss in terms of $\pi$. Therefore, we can derive the estimation of gradients for the expected loss:
\begin{align}\label{expected grad}
\begin{split}
        \frac{\partial \E_{x'\sim\pi}\mathcal{L}(f(x'),y)}{\partial \pi}\approx\frac{\partial}{\partial \pi}\E_{g}\mathcal{L}(f(x'(\pi, g)),y)\\
    =\E_{g}\left[\frac{\partial \mathcal{L}}{\partial x'}\frac{\partial x'}{\partial \pi}\right]\approx \frac{1}{n_g}\sum_{i=1}^{n_g}\left[\frac{\partial \mathcal{L}}{\partial x'}\frac{\partial x'(\pi, g_i)}{\partial \pi}\right]
\end{split}
\end{align}
where $n_g$ is the number of i.i.d samples from $g$. In Eq.(\ref{expected grad}), the first approximation is from the reparameterization of a sample $x'$; the second equality comes from exchanging the order of expectation and derivative, and the third approximation is to approximate the expectation of gradients by calculating the average of gradients.
Finally, we derive the practical solution to solve Eq.(\ref{final optimization}), by leveraging the gradient ascent algorithm, such as~\cite{DBLP:conf/iclr/MadryMSTV18}. In Algorithm~\ref{PCAA}, we provide the details of our proposed attack method. Specifically, during each iteration, we first estimate the gradient of expected loss (line 3), and then update the unnormalized distribution $\pi$ by gradient ascent (line 4). Finally, we clip $\pi$ back to its domain $(0,C]^d$ (line 5).

\begin{algorithm}[h]
\begin{algorithmic}[1]
\setstretch{1.3}
\INPUT Data $\mathcal{D}$, budget $\epsilon$, number of samples $n_g$, penalty coefficient $\lambda$, max iteration $I$, learning rate $\gamma$\\
\OUTPUT Adversarial Distribution $\pi$
\STATE Initialize distribution $\pi^0$
\FOR{$t\le I$}
\STATE Estimate expected gradient using Eq.\ref{expected grad}:  \\
$\nabla_{\pi}\E_{\pi}\mathcal{L}\approx \frac{1}{n_g}\sum_{i=1}^{n_g}\left[\frac{\partial \mathcal{L}}{\partial x'}\frac{\partial x'(\pi^t, g_i)}{\partial \pi}\right]$ \\
\STATE Gradient ascent: \\
$\widetilde{\pi}^{t+1} = \pi^t + \gamma \left(\nabla_{\pi} \E_{\pi}\mathcal{L}-\lambda\nabla_{\pi}[\mathcal{L}_{CE}(\pi_t, x)-\zeta]^{+}\right)$ \\
\STATE Clip to $(0,C]^d$: $\pi^{t+1}=\max(\widetilde{\pi}^{t+1}, C)$
\ENDFOR
\caption{Probabilistic Categorical Adversarial Attack}
\label{PCAA}
\end{algorithmic}
\end{algorithm}

\subsection{Time Complexity Analysis}\label{sec:time}
In this subsection, we compare the time complexity of \method with four representative attack methods~\citep{DBLP:conf/mlsys/LeiWCDDW19, DBLP:journals/jmlr/YangCHWJ20}. Notably, they are existing search-based methods to find adversarial examples for categorical data. Each of them consists of 2 stages: (1) the first stage is to search the top-K features that are most influential to the model output, which is determined by either manipulating the features and checking the loss change \textit{(loss-guided)} or the gradient scale \textit{(gradient-guided)}; (2) the second stage applies either a \textit{brutal search} or a \textit{greedy search} to find the optimal perturbation on the selected features. In Table~\ref{table: time complexity}, we summarize the main stages for different attack methods, and we name them as Search Attack \textbf{(SA)}, Greedy Attack  \textbf{(GA)}, Gradient-guided Search Attack  \textbf{(GSA)} and Gradient-guided Greedy Attack  \textbf{(GGA)}. More details can be found in Appendix \ref{append: time complex}. In Table~\ref{table: time complexity}, we assume that the whole dataset has $N$ data points, each data point has $n$ features, each feature has $d$ categories and the budget of the allowed perturbation is $\epsilon$. In the following time complexity analysis, one feedforward / backpropagate step is considered as one computational unit. Results are summarized in Table \ref{table: time complexity} where $C_1$ is a constant related to the number of samples $n_g$ and max iteration $I$ shown in Algorithm~\ref{PCAA}, and detailed time complexity analysis can be found in Appendix \ref{append: time complex}
\begin{table}[h!]
\centering
\caption{Time complexity analysis.}
\label{table: time complexity}
\resizebox{0.45\textwidth}{!}
{
\begin{tabular}{c|c|c| c }
\hline \hline
\textit{Method} & \textit{Stage 1} & \textit{Stage 2}& \textit{Time complexity}\\
\hline
SA & loss-guided & brutal & $N\cdot\mathcal{O}(nd+d^{\epsilon})$\\
\hline
GA & loss-guided & greedy & $N\cdot\mathcal{O}(nd+\epsilon d)$\\
\hline
GSA & gradient-guided & brutal &  $N\cdot\mathcal{O}(1+d^{\epsilon})$\\
\hline
GGA & gradient-guided&greedy &  $N\cdot\mathcal{O}(1+\epsilon d)$\\
\hline
PCAA &-&- &$C_1N\cdot \mathcal{O}(1)$\\
\hline
\hline
\end{tabular}}
\end{table}

From the analysis above, SA and GSA suffer from the exponential increase of time complexity when the number of feature categories $d$ and budget size $\epsilon$ is increasing. GA and GGA accelerate the algorithm and achieve better time efficiency. However, they can sacrifice the performance as they greatly narrow down the search space (see Table~\ref{table: attack} in Section~\ref{sec:exp_attack}). In Section~\ref{sec:exp_attack}, we further empirically show that \method can achieve significantly better optimality than GA and GGA, as well as  significantly lower computational cost than SA and GSA. Thanks to the advantage in efficiency and optimality, \method can fast generate strong adversarial examples. Moreover, because \method is a gradient-based method, the adversarial examples can be generated by batches. As a result, it can be easily incorporated into powerful defenses which are originally designed for continuous data. In the following section, we will present \defense, as an example to show \method's potential to be applied in popular adversarial defenses such as adversarial training and TRADES.
\section{Probabilistic Adversarial Training}

In this section, we provide an exemplar case to transfer one representative defense, PGD adversarial training~\cite{DBLP:conf/iclr/MadryMSTV18} to categorical defense. Note that we also extend another effective defense TRADES \cite{DBLP:journals/corr/abs-1901-08573} and the detail is shown in Appendix \ref{sec:trades}. It is also worth mentioning that other types of defenses such as certified defenses~\citep{cohen2019certified} also have the potential to be transferred to the categorical data via \method and we leave this exploration as future investigations. 

Based on PGD adversarial training for continuous data, we propose Probabilistic Adversarial Training (\defense) based on \method to train robust models for categorical data. Recalling the formulation of \method in Eq.(\ref{final optimization}), and denoting the parameters for classifier $f$ as $\theta$, the training objective for \defense is formulated as:

\vspace{-0.8cm}
\small{
\begin{equation*}\label{adv obj}\hspace{-0.3cm}
\min_{\theta}\left[\max_{\pi}\E_{x'\sim\pi} \left(\mathcal{L}(f(x';\theta),y)
-\lambda \left[ \sum_{i\in |n|}\mathcal{L}_{CE}(x_i, \pi_i)-\zeta\right]^{+}\right)\right]
\end{equation*}
}
\normalsize
\noindent Since our objective involves a penalty coefficient, we adopt the strategy in \cite{DBLP:conf/iclr/YurochkinBS20} to update $\lambda$ during training. Specifically, we adaptively choose $\lambda$ according to $\mathcal{L}_{CE}(x,\pi)-\zeta$ from the last iteration: when the value is large, we increase $\lambda$ to strengthen the constraints and vice versa. The implementation of \defense is illustrated in Algorithm~\ref{PAdvT}. Specifically, it first initializes model parameters (line 1); then during each iteration (from line 2 to line 9), it samples a mini-batch of data (line 3) and obtains an adversarial distribution for each data point through Algorithm~\ref{PCAA} (line 4 to 5), afterward $n_{adv}$ adversarial examples are sampled (line 6)  and used to update $\theta$ through Adam \cite{DBLP:journals/corr/KingmaB14} (line 8) and penalty coefficient $\lambda$ (line 9). The process will continue until the training process converges.

\begin{algorithm}[h]
\begin{algorithmic}[1]
\setstretch{1.1}
\INPUT data $\mathcal{D}$, parameters of clean model $\theta$, budget $\epsilon$, parameters of Algorithm~\ref{PCAA}, $n_{adv}$, initial penalty coefficient $\lambda^0$, penalty coefficient step size $\alpha$, parameters of Adam optimizer, number of iterations $I$\\
\OUTPUT parameters $\theta$ of the robust model\\
\STATE Initialize the network with a pre-trained robust model \\
\REPEAT
\STATE Sample mini-batch $B=\{x^1,...,x^m\}$ 
\FOR{$i=1,...,m~$ (in parallel)}
\STATE Apply Algorithm~(\ref{PCAA}) to $x^i$ to obtain adversarial distribution $\pi^i$\\
\STATE Sample $n_{adv}$ examples $\{x'^i_{1},...,X'^i_{n_{adv}}\}$ from $\pi^i$ using Gumbel Softmax
\ENDFOR
\STATE Update $\theta$ to minimize the average adversarial loss\\
\STATE \small{$\lambda= (\lambda-\alpha(\zeta-\frac{1}{m}\sum_{i\in [m]}\sum_{j\in[n]}\mathcal{L}_\text{CE}(x^i_j,\pi^i_j)))^+ $}\\
\UNTIL {Training converged}
\caption{Probabilistic Adversarial Training (\defense)}
\label{PAdvT}
\end{algorithmic}
\end{algorithm}

\normalsize
\section{Experiment}\label{experiment}

In this section, we conduct experiments to validate the effectiveness and efficiency of \method and \defense. In Section~\ref{sec:exp_attack}, we demonstrate that \method achieves a better balance between attack success rate and time efficiency. In Section~\ref{sec:exp_defense}, we empirically validate that \defense achieves good  robustness against categorical attacks. Our code is available at \url{https://anonymous.4open.science/r/categorical-attack-0B9B}.

\subsection{Categorical Adversarial Attacks}\label{sec:exp_attack}





\begin{table*}[h!]\label{pcaa ips}
\centering
\caption{Attacking performance on IPS, AG's news, and Splice datasets. ``SR." represents the attack success rate; ``T." denotes the average running time in seconds; and "-" indicates the running time over 10 hours. Each result runs 5 times, $95\%$ confidence intervals are shown.}
\resizebox{1\textwidth}{!}
{
\begin{tabular}{c|c| cc|cc |cc| cc | cc}
\hline
\hline
 \multirow{2}{*}{\textbf{Dataset}} & \textbf{Attack} &   \multicolumn{2}{c|}{\textbf{$\epsilon=1$}}
 & \multicolumn{2}{c|}{\textbf{$\epsilon=2$}}
  & \multicolumn{2}{c|}{\textbf{$\epsilon=3$}}
 & \multicolumn{2}{c|}{\textbf{$\epsilon=4$}}
  & \multicolumn{2}{c}{\textbf{$\epsilon=5$}}\\
  & \textbf{Method} &SR.($\uparrow$) &T.($\downarrow$) &SR.($\uparrow$) &T.($\downarrow$) &SR.($\uparrow$) &T.($\downarrow$) &SR.($\uparrow$) &T.($\downarrow$) &SR.($\uparrow$) &T.($\downarrow$)\\
\hline
\multirow{5}{*}{IPS} & \textbf{SA}& 66.11$\pm$0.03 & 38.5& \textbf{81.24$\pm$0.01} & 2028 & $-$&$-$&	$-$ & $-$ & $-$&$-$\\
&\textbf{GA} & 66.11$\pm$0.03 & 35.4 & 71.32$\pm$0.02 & 38.1 & 79.44$\pm$0.06&39.9&	85.28$\pm$0.07&41.5&	91.15$\pm$0.11&43.2\\
&\textbf{GSA}&41.53$\pm$0.04&\textbf{2.06}&	75.47$\pm$0.03&1022&	$-$&$-$& $-$&$-$& $-$&$-$ \\
&\textbf{GGA} &41.53$\pm$0.04 & \textbf{2.06}&	63.72$\pm$0.06 & \textbf{2.01} &	70.76$\pm$0.05& \textbf{2.94} & 75.89$\pm$0.08&\textbf{3.74}&	82.43$\pm$0.10&\textbf{4.56}\\
&\textbf{PCAA}&\textbf{67.56$\pm$0.05}&14.04&	80.51$\pm$0.06&13.75&	\textbf{88.37$\pm$0.07} & 13.02& \textbf{93.67$\pm$0.04}&12.21&	\textbf{96.63$\pm$0.12}&14.59 \\
\hline

\multirow{5}{*}{AG}
&\textbf{SA}&41.22$\pm$0.01&15.3&	\textbf{67.38$\pm$0.02}&21.9&	75.87$\pm$0.04 & 356&	83.10$\pm$0.02&19160	& $-$ &$-$\\
&\textbf{GA}&41.22$\pm$0.01&15.3&	60.71$\pm$0.05&15.4&	66.33$\pm$0.04&15.5&	74.47$\pm$0.07&15.7&	86.63$\pm$0.12&15.9 \\
&\textbf{GSA}&32.39$\pm$0.03 & \textbf{0.352}&	59.21$\pm$0.02&3.24&	67.79$\pm$0.03&151&	79.22$\pm$0.05&7551&	$-$&$-$\\
&\textbf{GGA}&32.39$\pm$0.03 & \textbf{0.352}&	41.29$\pm$0.07&\textbf{0.393}&	56.11$\pm$0.06& \textbf{0.511} &	67.53$\pm$0.10& \textbf{0.613} & 72.28$\pm$0.07&\textbf{0.856} \\
&\textbf{PCAA}&\textbf{46.31$\pm$0.07}&16.03&	67.27$\pm$0.08&15.79&	\textbf{76.71$\pm$0.06}&19.21&	\textbf{84.65$\pm$0.09}&17.83&	\textbf{90.21$\pm$0.11}&17.35 \\
\hline

\multirow{5}{*}{Splice} &\textbf{SA}&\textbf{72.11$\pm$0.01}&0.905&	79.02$\pm$0.02&1.02&	\textbf{86.59$\pm$0.01} & 1.36 &	90.11$\pm$0.02 & 2.73 & 92.58$\pm$0.03 & 8.29\\
&\textbf{GA}& \textbf{72.11$\pm$0.01}&0.905&	74.42$\pm$0.03&0.911&	78.18$\pm$0.06&0.915&	80.61$\pm$0.04&0.922&	83.74$\pm$0.05&0.928\\
&\textbf{GSA}&61.71$\pm$0.02	&\textbf{0.028}&68.28$\pm$0.03&0.083&	72.82$\pm$0.02&0.251&	77.11$\pm$0.04&0.917&	82.53$\pm$0.04&3.56 \\
&\textbf{GGA}& 61.71$\pm$0.02&\textbf{0.028}& 65.26$\pm$0.08 & \textbf{0.031} & 70.31$\pm$0.05 & \textbf{0.0337} & 74.84$\pm$0.07 & \textbf{0.035} & 80.49$\pm$0.08 & \textbf{0.037}\\
&\textbf{PCAA}&72.05$\pm$0.03&3.27	&\textbf{79.33$\pm$0.06}&2.82&	86.12$\pm$0.07&3.18&	\textbf{90.33$\pm$0.06}&2.56&	\textbf{92.90$\pm$0.08}&3.02\\
\hline
\hline
\end{tabular}}
\label{table: attack}
\end{table*}

\begin{figure*}[t]
\centering
\begin{subfigure}[b]{0.3\textwidth}
        \centering
        \includegraphics[height=1.3in]{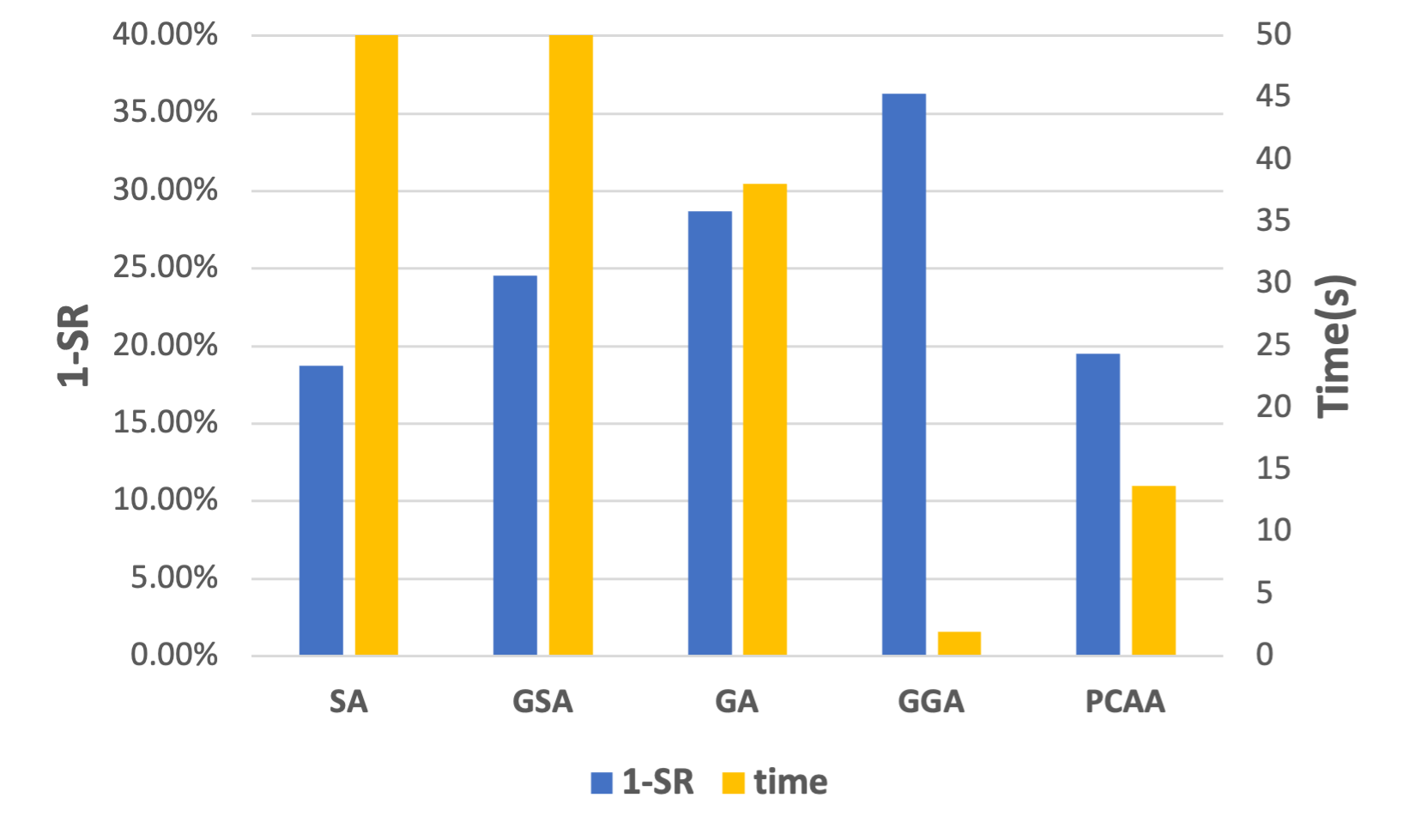}
        \caption{IPS with $\epsilon = 2$}
    \end{subfigure}~
    \begin{subfigure}[b]{0.3\textwidth}
        \centering
        \includegraphics[height=1.3in]{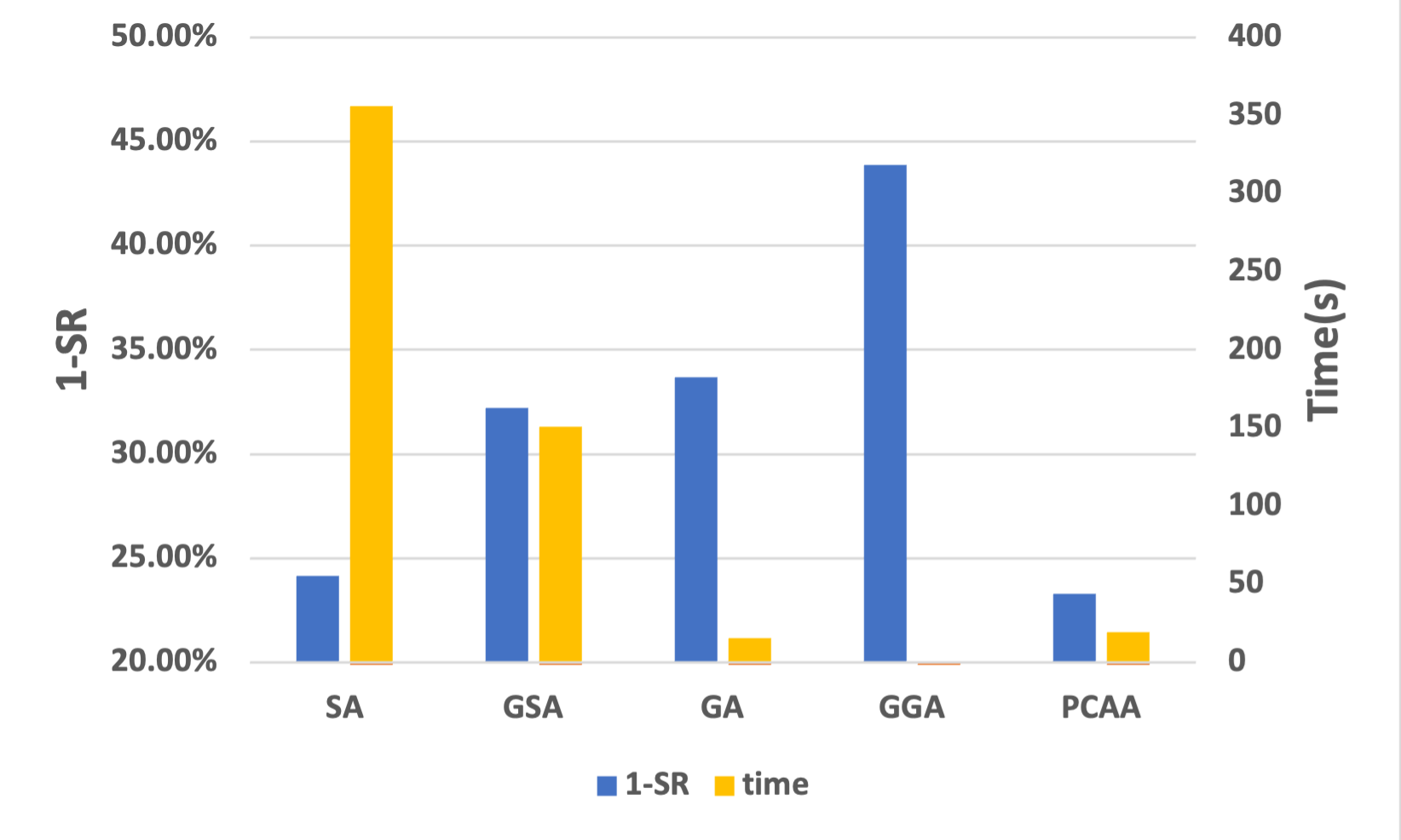}
        \caption{AG's news with $\epsilon = 3$}
    \end{subfigure}~
    \begin{subfigure}[b]{0.3\textwidth}
        \centering
        \includegraphics[height=1.3in]{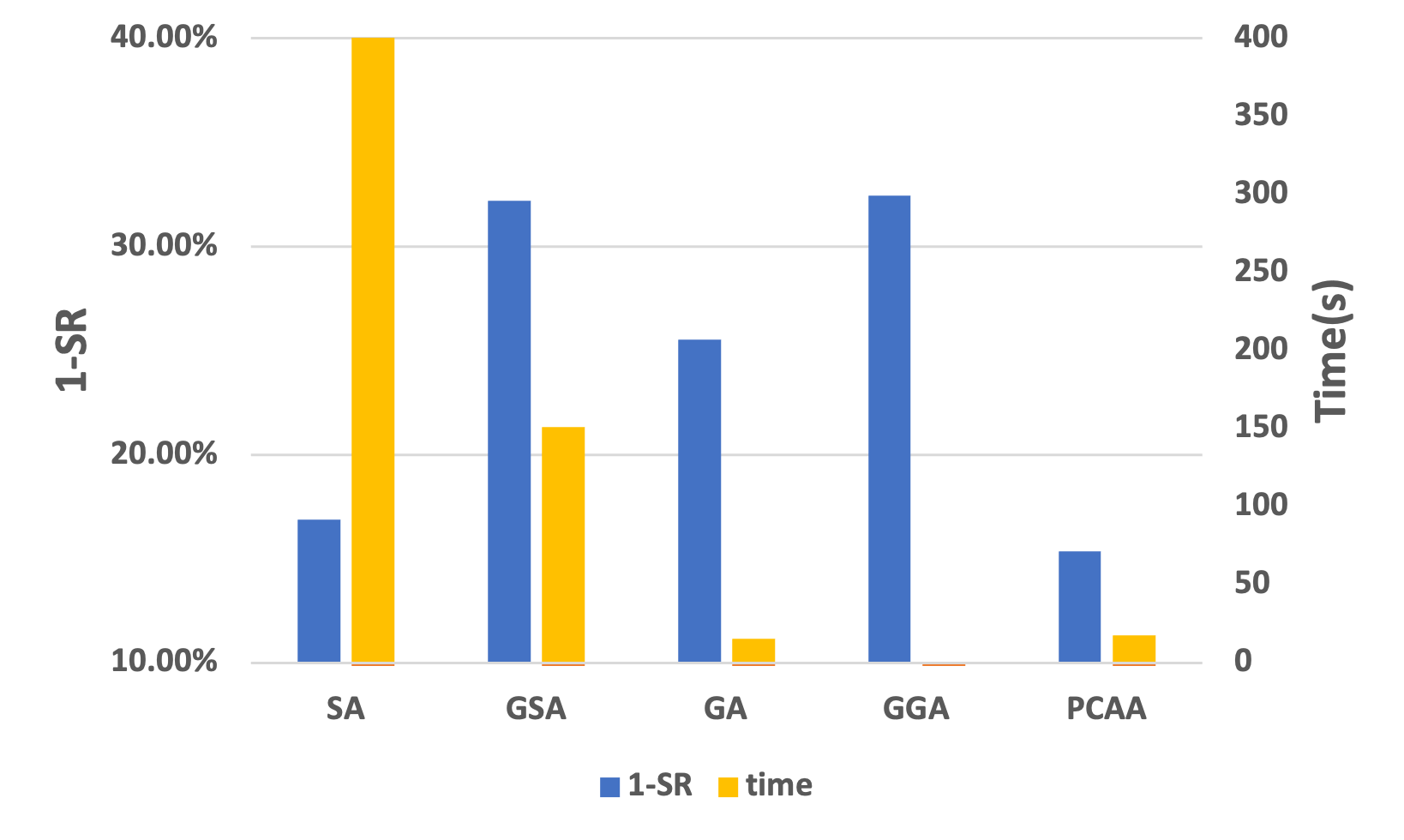}
        \caption{AG's news with $\epsilon = 4$}
    \end{subfigure}
    \caption{An illustration of attack success rate and time efficiency trade-offs. The blue bar represents $1-SR$ under different attacks while the yellow bar denotes the average running time. For both metrics, smaller values indicate stronger attacks. 
    }
    \label{fig: tradeoffs}
\end{figure*}


{\bf Experimental Setup.} In this evaluation, we focus on three categorical datasets for various applications.\textbf{(1) Intrusion Prevention System (IPS)~\citep{DBLP:conf/kdd/WangHBSML020}.}  IPS dataset has 242,467 instances, with each input consisting of 20 features and each feature has 1,103 categorical values. The output space has three labels. A standard LSTM based classifier\citep{DBLP:conf/iclr/BaoHZS022} is trained for IPS dataset.   \textbf{(2) AG’s News corpus.} This dataset consists of titles and description fields of news articles. The tokens of each sentence correspond to the categorical features, and the substitution set (of size 70) corresponds to the categorical values. A character-based CNN\citep{zhang2015character} is trained on this dataset.  \textbf{(3) Splice-junction Gene Sequences (Splice)~\citep{noordewier1990training}.} Splice dataset has 3,190 instances. Each one is a gene fragment of 60 features with 5 categorical values. The output space has three labels and the model is LSTM.

{\bf Baseline Attacks.} 
We compare PCAA with the following search-based attacks including SA, GA, GSA and GGA, which are discussed in Section~\ref{sec:time}. The details of these attacks can also be found in Appendix \ref{append: time complex}. 

\textbf{Implementation details.} For each dataset, we evaluate the performance in terms of the attack success rate (SR.) and the average running time (T.) under various budget sizes $\epsilon$ ranging from $1$ to $5$. 
Remind that in PCAA in Eq.(\ref{final optimization}), the threshold $\zeta$ significantly influences the effectiveness of our method. Therefore, in Table~\ref{table: attack}, we iteratively conduct \method with different choices of $\zeta$ from a pre-defined set. Given each $\zeta$, we make 100 samplings from the probabilistic distribution.
In this process, once a successful adversarial example (satisfying the $l_0$ budget constraint) is generated, we claim it to be a successful attack. 

\textbf{Performance Comparison.} The experimental results on IPS, AG's news, and Splice datasets are demonstrated in Table~\ref{table: attack}. The results clearly show that our \method reaches the best balance between optimality and efficiency. 

\textbf{(1) PCAA vs.~SA / GSA.} Both SA and GSA apply brutal search and this leads to terrible efficiency in practice, especially when the budget and the dimension of input space are large (in datasets such as IPS and AG). On IPS dataset, in which each data point consists of more than 1,000 categorical features, when the budget is more than 2, SA and GSA become infeasible leading to their incapability of either practical attack or defense. Compared to SA or GSA, the time complexity of PCAA does not increase with the increase of perturbation budget $\epsilon$. Moreover, PCAA always has a better (or at least a similar) successful attack rate than SA or GSA.

\textbf{(2) PCAA vs.~GA / GGA.} These methods accelerate the search process(second stage) by leveraging greedy algorithms. They achieve good efficiency on all 3 datasets, especially on datasets with a small number of categorical features such as the Splice dataset. However, they sacrifice the performance significantly and usually obtain a success rate of 10\% less than other methods. The lack of optimality prevents these methods from generating practical attacks and further usage for defenses. Our \method does not have this concern as it outperforms them by significant margins, e.g., over 12\% higher than GA/GGA in success rate, while still having an acceptable running time. 

To further demonstrate that \method achieves a better balance, we visualize results on some datasets with different budgets based on Table~\ref{table: attack} in Figure~\ref{fig: tradeoffs}, where we plot \textbf{($1 - \text{success rate}$)} and \textbf{time} for each method. From the figure, all baseline methods have high levels on at least one metric, while \method can keep both metrics under low levels. 
Therefore, \method can overcome the drawbacks of baseline methods and achieve better efficiency vs.~optimality trade-off.

\begin{figure*}
    \centering
    \includegraphics[width = 0.9\textwidth]{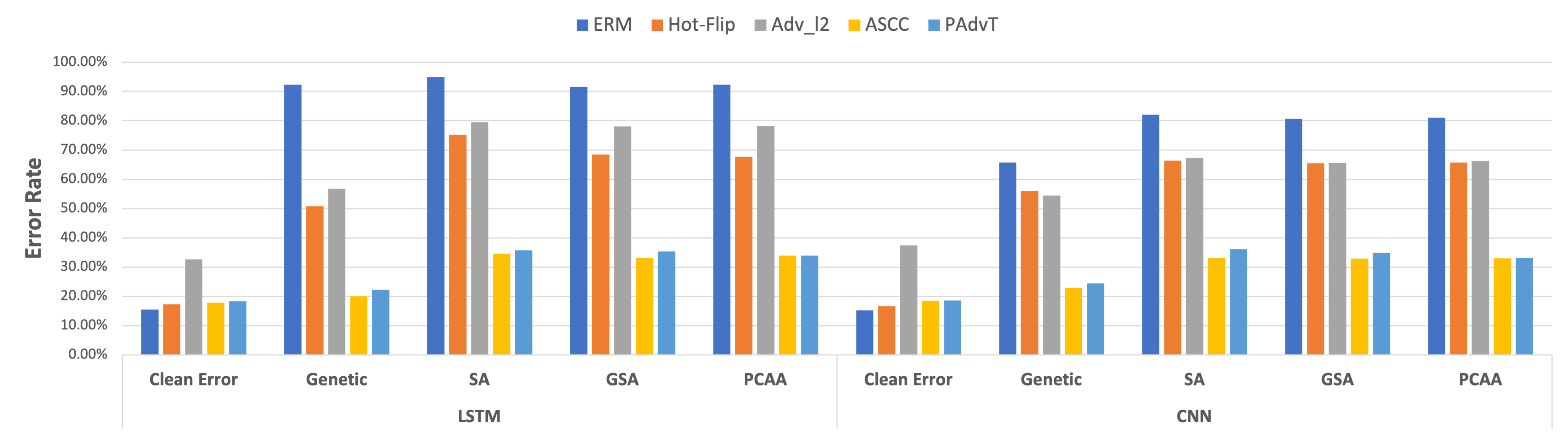}
    \vspace{-0.2in}
    \caption{PAdvT and baseline defense performance under different attacks on IMDB dataset.}
    \label{table2_defense}
\end{figure*}

\begin{figure}
    \centering
    \includegraphics[width = 0.4\textwidth]{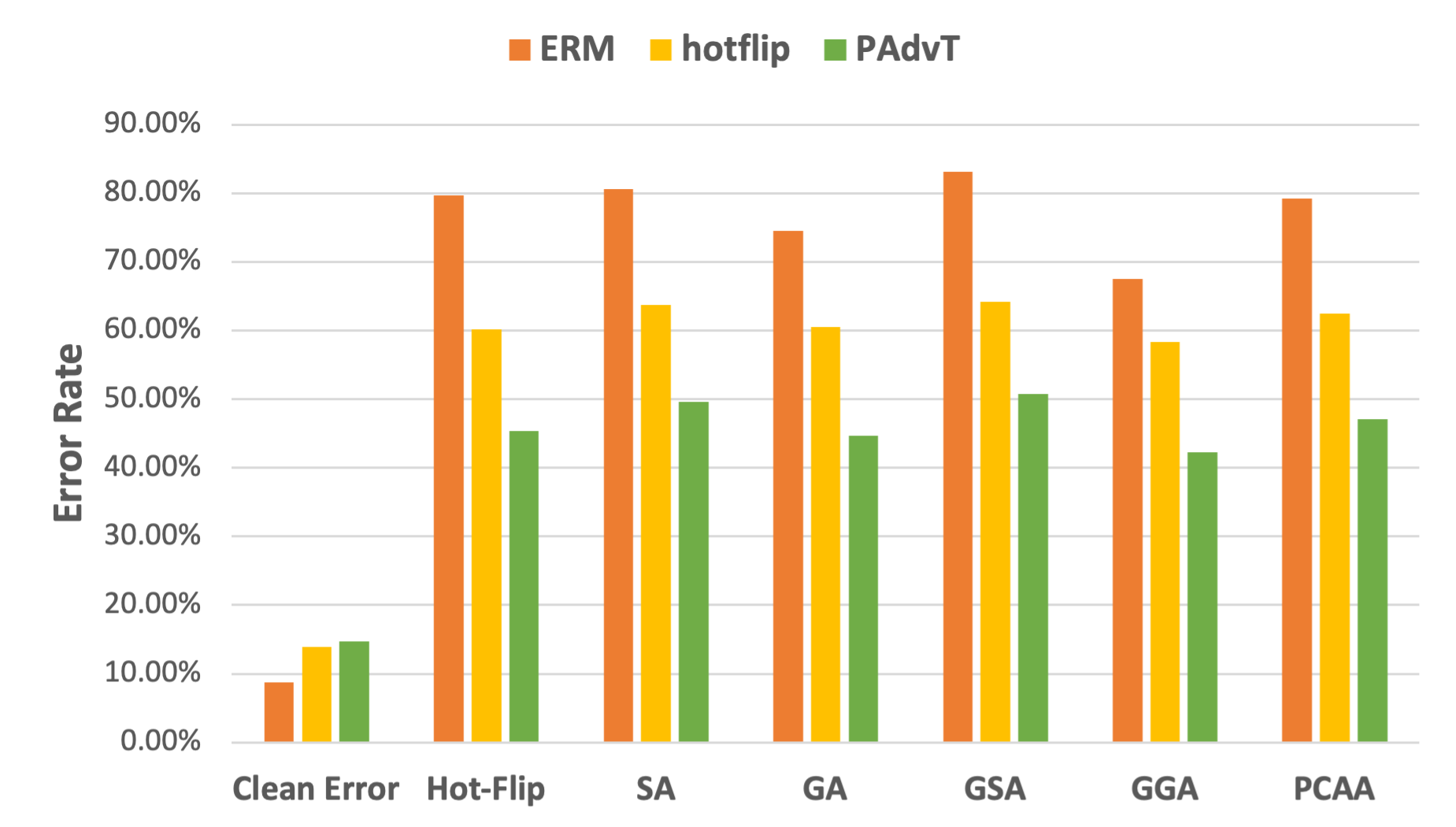}
    \caption{PAdvT and baseline defense performance under different attacks on AG's news dataset.}
    \label{table3_defense}
\end{figure}

\subsection{Categorical Adversarial   Defenses}\label{sec:exp_defense}

\textbf{Experimental Setup.} For the defense evaluation, we focus on two datasets, AG's News Corpus and IMDB. It is because  IPS and Splice have too few samples (no more than 1,000 for each dataset). In particular: \textbf{(1) AG’s News corpus.} is the same dataset used in the attack evaluation and the model is also a character-based CNN. As character swapping does not require embeddings for each character, we can directly apply attacking methods on input space. Therefore, the robustness of the defense models are evaluated using six attacks, i.e., Hot-Flip~\citep{ebrahimi2017hotflip}, SA, GA, GSA, GGA, and PCAA. \textbf{(2) IMDB reviews dataset}~\citep{DBLP:conf/acl/MaasDPHNP11}. Under this dataset, we focus on a word-level classification task and we study two model architectures, namely Bi-LSTM and CNN, trained for prediction. To evaluate the robustness, four attacks are deployed, including a genetic attack~\citep{DBLP:conf/emnlp/AlzantotSEHSC18} (which is an attack method proposed to generate adversarial examples in embedding space), as well as SA, GSA, and \method. Note that for text data, we also consider preserving semantic meanings and grammatical correctness during defenses, and only perturb words with synonyms and correct grammatical forms.

\textbf{Baseline Defenses.} We compare our defense method \defense with the following existing baseline defenses:

\begin{itemize}[leftmargin=0.3in]
\setlength\itemsep{0em}
    \item \textbf{Standard training}. It minimizes the average cross-entropy loss on clean input.
    \item \textbf{Hot-flip}~\citep{ebrahimi2017hotflip}. It uses the gradient with respect to the one-hot input representation to find out which individual feature under perturbation has the highest estimated loss. It is initially proposed to model char flip in Char-CNN model, and we also apply it to word-level substitution, as in \citep{DBLP:conf/iclr/DongLJ021}.
    \item \textbf{Adv $l_2$-ball}~\citep{miyato2017adversarial}. It uses an $l_2$ PGD adversarial attack inside the word embedding space for adversarial training. 
    \item \textbf{ASCC-Defense}~\citep{DBLP:conf/iclr/DongLJ021}. A state-of-the-art defense method in text classification. It uses the worst perturbation over a sparse convex hull in the word embedding space for adversarial training. 
\end{itemize}

\textbf{Legible attacks on NLP.}  It is worth noting that when we apply our method to the text datasets IMDB, we consider the additional requirements of maintaining semantic meaning and grammatical correctness for NLP tasks. Following existing textual data attacking methods \cite{DBLP:conf/iclr/DongLJ021}, we only switch some words with their synonyms while keeping correct grammatical forms and perturb each word separately. Specifically, we follow the same way of \cite{DBLP:conf/iclr/DongLJ021} which constructs the feasible perturbation space $\mathcal{S}(x)$ (see Eq.(\ref{deter form})) to maintain the semantic meaning and avoid grammatical errors, and apply the perturbation space into our framework. In Appendix \ref{texts}, we provide some real examples to show how words are replaced with their synonyms while keeping correct grammatic forms.  We also provide adversarial texts for human evaluation to further confirm that this strategy meets the aforementioned requirements.

\textbf{Performance Comparison.} The experimental results on AG's news and IMDB are shown respectively in Fig.~\ref{table3_defense} and Fig.~\ref{table2_defense}. The Y-axis represents the error rate and X-axis represents different attacks where each bar inside the group of an attack denotes one defense method. On AG's news dataset, our defense method achieves leading robustness on Char-CNN over all attacks with significant margins, surpassing Hot-Flip-defense by $10\%$. On the IMDB dataset, and we have similar observations to these on AG's news dataset. Our \defense shows competitive adversarial robustness as ASCC defense. Notably, ASCC is a defense method that conducts adversarial training on word-embedding space. It relies on the key assumption that similar words have a close distance in embedding space. However, our method does not rely on this assumption, which may result in the performance being competitive (slightly worse) than ASCC. For all other defenses, \defense outperforms them across different architectures significantly. Note that \defense is based on PGD adversarial training.  We also adapt TRADES to categorical data based on \method and details are shown in Appendix~\ref{sec:trades}.

\begin{table*}[t!]
\centering
\caption{\small Ablation study: impact of the budget regularization term $\zeta$ on PAdvT}
\resizebox{0.75\textwidth}{!}
{
\begin{tabular}{c| cc|cc |cc| cc | cc}
\hline\hline
 &   \multicolumn{2}{c|}{\textbf{Clean Err}}
 & \multicolumn{2}{c|}{\textbf{Genetic SR}}
  & \multicolumn{2}{c|}{\textbf{SA SR}}
 & \multicolumn{2}{c|}{\textbf{Gradient Search SR}}
  & \multicolumn{2}{c}{\textbf{PCAA SR}}\\
  &LSTM &CNN &LSTM &CNN &LSTM &CNN  &LSTM &CNN &LSTM &CNN\\
\hline
\textbf{$\zeta=0.1$}&16.1&16.9&	37.5&29.6&	41.3&43.5&	39.5&41.4&	40.2&42.2\\
\textbf{$\zeta=0.2$}&17.2&17.3&	34.1&32.7&	39.9&41.3&	38.6&39.7&	39.1&40.6\\
\textbf{$\zeta=0.32$}&17.9&18.1&	30.1&31.4&	38.7&38.3&	38.5&37.6&	38.8&37.8 \\
\textbf{$\zeta=0.4$}&18.4&18.6&	26.3&28.5&	37.8&36.2&	36.4&34.9&	36.9&35.7\\
\hline
\hline
\end{tabular}}
\label{ablation}
\end{table*}

\begin{figure}[t!]
    \centering
     \includegraphics[width = 0.5\textwidth]{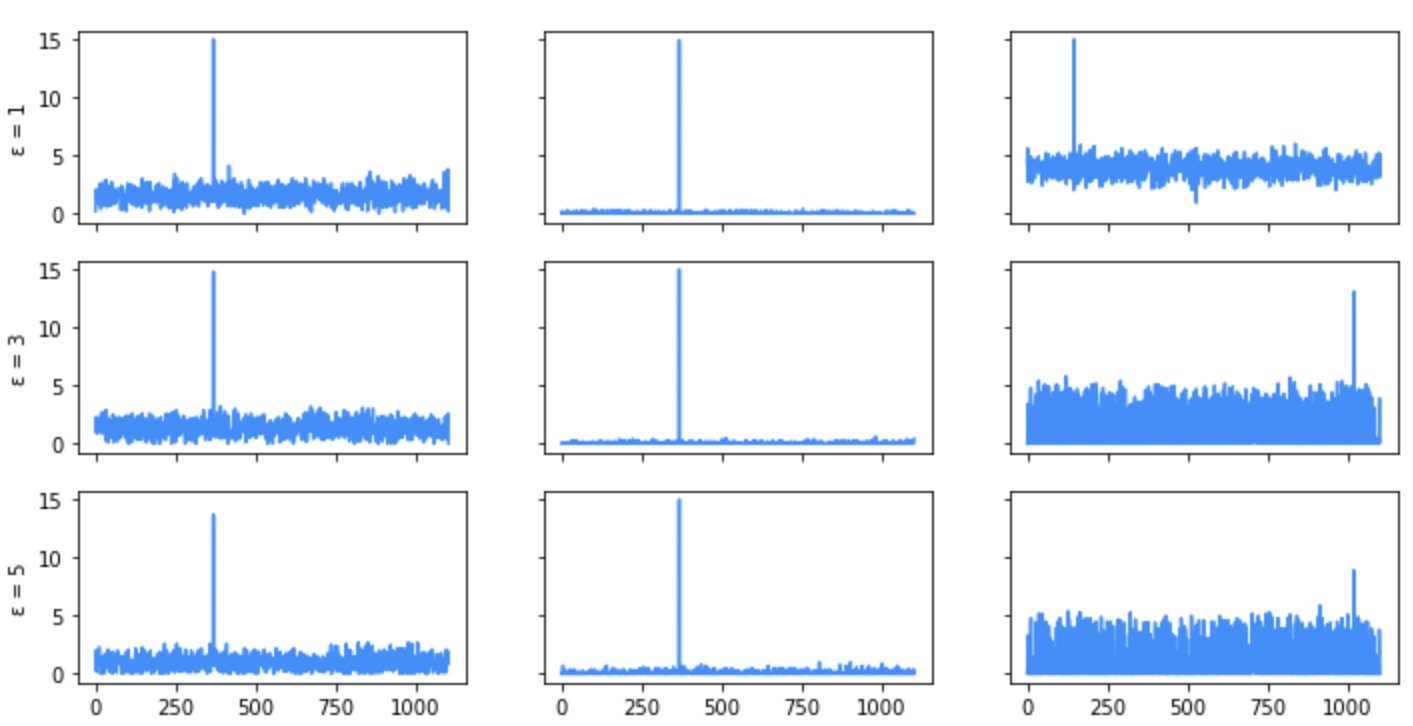}
    \caption{Visualization of Optimized Categorical Distribution for Various Features (IPS)}
    \label{sparsity illu}
\end{figure}

\subsection{Ablation Study}

\textbf{Concentration of \method.} To further understand the behavior of our attack algorithm, in this subsection, we ask the question: \textit{what is the variance of our optimized probability distribution $\pi^*$ (from solving Eq.(\ref{prob cons})?}
Intuitively, we desire the distribution $\pi$ to have a smaller variance, so that we don't need too many times sampling to obtain the optimal adversarial examples. To confirm this point, 
we conduct an ablation study based on an experiment on IPS dataset to visualize the distribution $\pi$, which is optimized via \method under various budget sizes. In Fig.~\ref{sparsity illu}, we choose three budget sizes $\epsilon=1,3,5$ and randomly choose 3 features to present the adversarial categorical distributions, where the Y-axis represents the magnitude of unnormalized probabilities for each level within the feature. Notably, in Fig.~\ref{sparsity illu}, the left and middle two columns correspond to the feature distribution where the most probable category is the same as original category, and the right column are the feature that where the most probable category is different. From the figure, we can see that for all features, there exists one category with a much higher probability compared to other categories. This fact indicates that during the sampling process of \method, the samples are highly likely to have the same category for a certain feature. As a result, we confirm that our sampled adversarial examples are well-concentrated.

\textbf{The Impact of $\zeta$ on \defense.} In our training objective in Eq.(\ref{adv obj}), $\zeta$ controls the budget size used for adversarial training and possibly affects the robustness of the model. We conduct an ablation study on the IMDB dataset to understand the impact of $\zeta$. The results are demonstrated in Table~\ref{ablation}. When $\zeta$ increases, the success rates of all attacks decrease, meaning that the robustness of the models is enhanced. However, large $\zeta$ will decrease the model accuracy. Thus, $\zeta$ controls the balance between the accuracy and the robustness of the model. When $\zeta=0.4$, our algorithm reaches a good balance between accuracy and robustness.

\section{Conclusion}
In this paper, we propose a novel probabilistic framework, PCAA, to bridge the gap between categorical data and continuous data, which allows us to easily adapt gradient-based attacking methods in continuous data to categorical data. Our framework significantly improves the optimality-efficiency trade-off compared with search-based methods and shows promising empirical performance across different datasets. Furthermore, we adapt defenses in continuous data to categorical data through the proposed framework and achieve better robustness. Our future work will pursue transferring other advanced methods designed for the continuous domain, such as certified defenses~\citep{cohen2019certified}, to the categorical domain.

\section{Acknowledgements}

Han Xu, Pengfei He, Jie Ren and Jiliang Tang are supported by the National Science Foundation (NSF) under grant numbers CNS1815636, IIS1845081, IIS1928278, IIS1955285, IIS2212032, IIS2212144, IOS2107215, and IOS2035472, the Army Research Office (ARO) under grant number W911NF-21-1-0198, the Home Depot, Cisco Systems Inc, Amazon Faculty Award, and SNAP. Zitao Liu was supported in part by National Key R\&D Program of China, under Grant No. 2022YFC3303600; and in part by Key Laboratory of Smart Education of Guangdong Higher Education Institutes, Jinan University (2022LSYS003).

\balance
\bibliography{example_paper}
\bibliographystyle{icml2023}

\appendix
\onecolumn
\section{Appendix}

\subsection{Detailed computation for time complexity}\label{append: time complex}

In this section, we provide the detailed computation of time complexity analysis mentioned in Section \ref{sec:time}. Recall the assumptions that the whole dataset has $N$ data points, each data point has $n$ features, each feature has $d$ categories and the budget of the allowed perturbation is $\epsilon$. In the following time complexity analysis, one feedforward / backpropagate step is considered as one computational unit.

\textbf{\method}. In \method, the time complexity is only from gradient ascent. Here, we assume that we sample $n_g$ times when estimating the expected gradient, and the maximum number of iterations is $I$ in Algorithm~\ref{PCAA}. We compute gradient $n_g$ times during one iteration, which consists of one feedforward and one backpropagate step. Thus, the time complexity is:
\begin{equation}
    N\cdot n_g \cdot \mathcal{O}(1) \cdot I=C_1N \cdot \mathcal{O}(1)
\end{equation}
where $C_1$ is some constant related to $n_g$ and $I$.

\textbf{Search Attack (SA)}. SA consists of two stages. The first stage involves traversing all features. For the $i_{th}$ feature, it replaces the original category with all other $d-1$ categories respectively, and records the change of the model loss for each category. The largest change is treated as the impact score for the $i_{th}$ feature. Then it selects the top $\epsilon$ features with the highest impact scores to perturb. In the second stage, it finds the combination with the greatest loss among all possible combinations of categories for selected features. Each loss calculation above involves one feedforward step and totally there are $nd+d^{\epsilon}$ loss calculations. Therefore, the time complexity for SA is
$$\label{tc gs}
N\cdot[\mathcal{O}(nd)+\mathcal{O}(d^{\epsilon})]=N\cdot\mathcal{O}(nd+d^{\epsilon})
$$
\textbf{Greedy attack (GA)}~\citep{DBLP:journals/jmlr/YangCHWJ20}. This method is a modified version of 
SA. The first stage is similar to that of SA, while the second stage searches for the best perturbation feature by feature via greedy search. For the $i_{th}$ selected feature, it replaces the original category with one that results in the largest loss and then searches the next selected feature until all selected features are traversed. Each loss calculation above involves one feedforward step and totally there are $nd+\epsilon d$ loss calculations. It has the complexity: $$N\cdot[\mathcal{O}(nd)+\mathcal{O}(\epsilon d)]=N\cdot\mathcal{O}(nd+\epsilon d)$$
\textbf{Gradient-guided SA (GSA)}~\citep{DBLP:conf/mlsys/LeiWCDDW19}. To determine which features to perturb, this method utilizes gradient information in the first stage. It computes the gradient of the loss function w.r.t the original input and treats the gradient of each feature as the impact score. Those $\epsilon$ features with the greatest impact scores are selected to be perturbed. In the second stage, it follows the same strategy as that of SA. The gradient calculation involves one feedforward and one backpropagate step, and the loss calculation involves one feedforward step per feature. Therefore the time complexity is: $$
N\cdot[\mathcal{O}(1)+\mathcal{O}(d^{\epsilon})]=N\cdot\mathcal{O}(1+d^{\epsilon})$$
\textbf{Gradient-guided GA (GGA)}. On the basis of GSA, it remains the same first stage and modifies the second stage by adopting the same strategy as that in the second stage of GA. Thus its time complexity is:
$$N\cdot[\mathcal{O}(1)+\mathcal{O}(\epsilon d)]=N\cdot\mathcal{O}(1+\epsilon d).$$

\subsection{Additional experimental results}

To better compare the performance of different defenses, we provide exact results (error rates) corresponding to Fig.~\ref{table3_defense} and Fig.~\ref{table2_defense} and show them in Table~\ref{Table: padvt IMDB} and Table~\ref{Table: padvt AG}, respectively. Specifically, we run each experiment 5 times and compute the $95\%$ confidence interval.

We also run PAdvT with a mixture of adversarial examples and clean samples on the IMDB dataset. Results are shown in Table \ref{table: mixture} where values represent error rates. It is noticeable that clean samples will slightly improve the clean performance and lead to a small decrease in robustness.

\begin{table*}[h!]
\centering
\caption{\small PAdvT and baseline defense performances under different attacks on IMDB dataset}
\label{Table: padvt IMDB}
\resizebox{1\textwidth}{!}
{
\begin{tabular}{c| cc|cc |cc| cc | cc}
\hline\hline
 &   \multicolumn{2}{c|}{\textbf{Clean Err}}
 & \multicolumn{2}{c|}{\textbf{Genetic SR}}
  & \multicolumn{2}{c|}{\textbf{GS SR}}
 & \multicolumn{2}{c|}{\textbf{Gradient Search SR}}
  & \multicolumn{2}{c}{\textbf{PCAA SR}}\\
  &LSTM &CNN &LSTM &CNN &LSTM &CNN  &LSTM &CNN &LSTM &CNN\\
\hline
\textbf{ERM}&15.50$\pm$0.005&15.23$\pm$0.007&	92.62$\pm$0.022&65.68$\pm$0.031&	94.86$\pm$0.030&82.02$\pm$0.049&	91.61$\pm$0.011&80.54$\pm$0.030&	92.26$\pm$0.087&81.42$\pm$0.073\\
\textbf{Hotflip}&17.37$\pm$0.049&16.57$\pm$0.051&	50.63$\pm$0.030&55.93$\pm$0.025&	75.24$\pm$0.015&66.35$\pm$0.012&	67.96$\pm$0.021&65.47$\pm$0.022&	68.08$\pm$0.069&65.90$\pm$0.074\\
\textbf{Adv\_l2}&32.53$\pm$0.034&37.38$\pm$0.043&	56.59$\pm$0.036&54.35$\pm$0.033&	79.69$\pm$0.030&67.00$\pm$0.034&	78.34$\pm$0.031&65.81$\pm$0.038&	78.58$\pm$0.068&66.23$\pm$0.077 \\
\textbf{ASCC}&17.76$\pm$0.036&18.37$\pm$0.030&	20.05$\pm$0.047&22.52$\pm$0.046&	34.67$\pm$0.047&33.28$\pm$0.045&	33.46$\pm$0.054&32.61$\pm$0.061&	33.97$\pm$0.101&33.01$\pm$0.082\\
\textbf{PAdvT}&18.57$\pm$0.033&18.85$\pm$0.049&	22.32$\pm$0.065&24.50$\pm$0.049&	37.30$\pm$0.063&36.36$\pm$0.061&	35.60$\pm$0.067&34.85$\pm$0.053&	33.90$\pm$0.104&33.20$\pm$0.093\\
\hline
\hline
\end{tabular}}
\end{table*}

\begin{table*}[h!]
\centering
\caption{\small PAdvT and baseline defense performances under different attacks on AG's news dataset}
\label{Table: padvt AG}
\resizebox{1\textwidth}{!}
{
\begin{tabular}{c| c|c |c| c | c|c|c}
\hline\hline
 &Clean Err & Hotflip & PCAA & GS & GGS & GA & GGA\\
\hline
\textbf{ERM}&8.70$\pm$0.009&	79.72$\pm$0.015&80.75$\pm$0.066&	83.09$\pm$0.015&79.28$\pm$0.010&	74.43$\pm$0.010&67.53$\pm$0.016\\
\textbf{Hotflip}&13.99$\pm$0.017&	60.07$\pm$0.013&63.47$\pm$0.057&	64.28$\pm$0.018&62.41$\pm$0.019&	60.51$\pm$0.017&58.33$\pm$0.013\\
\textbf{PAdvT}&14.62$\pm$0.028&45.38$\pm$0.037&	49.50$\pm$0.081&50.65$\pm$0.035&	47.14$\pm$0.041&44.71$\pm$0.040&	42.18$\pm$0.045\\
\hline
\hline
\end{tabular}}
\end{table*}

\begin{table*}[h!]
\centering
\caption{Comparison of PAdvT on IMDB dataset with/without mixture of clean samples.}
\label{table: mixture}
\begin{tabular}{c| c|c|c|c|c }
\hline\hline
& Clean Err& Genetic & GS & GGS & PCAA \\
\hline
\textbf{IMDB LSTM(mix)}&18.27&26.22&37.67&35.79&36.05\\
\hline
\textbf{IMDB LSTM}&18.57&	26.01&	37.3&	35.60&	35.46\\
\hline
\textbf{IMDB CNN(mix)}&18.69&28.51&36.47&34.96&35.72\\
\hline
\textbf{IMDB CNN}&18.85&	28.35&	36.36&	34.85&	35.47\\
\hline
\hline
\end{tabular}
\end{table*}

\subsection{A categorical defense based on TRADES}\label{sec:trades}
To better show the capability of our probabilistic framework, we apply another effective continuous defense method, TRADES~\cite{DBLP:journals/corr/abs-1901-08573}, on the categorical dataset AG's news through our framework, and the results are shown in Table \ref{table: trades}. In detail, we modify Algorithm~\ref{PAdvT} to implement this defense, and replace the average adversarial loss in line 8 with the TRADES loss on probabilistic distribution $\pi$, i.e
$$
\mathcal{L}(f(x,\theta),Y)+\max_{\pi}\E_{x'\sim\pi}\mathcal{L}(f(x',\theta), f(x,\theta))/\lambda
$$
According to the results, our framework can be easily leveraged for TRADES and achieves good performance, and can reduce error under different attacks. We conduct this defense with different choices of regularizer parameter $1/\lambda$, and the results show that the combination of \method and TRADES can achieve both high robustness and high accuracy on categorical data, indicating the capability of our framework.

\begin{table*}[t!]
\centering
\caption{\small PAdvT and TRADES on AG's news}
\resizebox{0.7\textwidth}{!}
{
\begin{tabular}{c| c|c |c| c | c|c|c}
\hline\hline
 &	clean error&	hotflip&	PCAA&	SA&	GSA&	GA&	GGA\\
\hline
\textbf{ERM}&8.70 & 79.72 & 80.75 & 83.09 &	79.28 & 74.43 & 67.53\\
\textbf{PAdvT}&14.62 & 45.38 & 49.50 & 50.65 & 	47.14 & 44.71 & 42.18\\
\textbf{TRADES($1/\lambda=1$)}&13.89&	46.21&	50.83&	51.32&	48.45&	45.98&	43.17 \\
\textbf{TRADES($1/\lambda=5$)}&14.31&	45.57&	49.76&	50.89&	47.66&	45.12&	42.31\\
\hline
\hline
\end{tabular}}
\label{table: trades}
\end{table*}

\subsection{Case studies on IMDB}\label{texts}

To better illustrate how our method can maintain original meanings and grammaticals when applied to NLP tasks, we provide some case studies on IMDB dataset. First, we present 2 cases including original texts and replacement for the perturbed words. The first case shows that theses words will be replaced with their synonyms, and the second case indicates that the grammatical form of replaced words will be consistent with the original words. Original words are marked in blue, while their replacements are listed in red.
\begin{itemize}
    \item \textbf{Case 1}. Synonyms.

    The cast is {\color{blue}excellent}({\color{red}\{admirable, distinguished, exquisite, finest, first-rate, good, magnificent ,Outstanding, skillful, sterling, superb, marvelous, best, attractive, great, exceptional, accomplished, fine, exemplary, first-class\}}), the acting good, the plot interesting, the evolvement
full of suspense, but it is hard to cram all those elements into a film that is barely 80
minutes long. If more time was taken to develop the plot and subplots, it would have
a much better effect. Another 30 minutes of substance would have made this a very
{\color{blue}good}({\color{red}\{acceptable, exceptional, favorable, great, right, marvelous, satisfying, superb ,valuable, wonderful, ace, bad, capital, nice, pleasing, excellent, positive, satisfactory, excellent, fine\}}) film rather than just a good one. 

\item \textbf{Case 2}. Grammatricals.

There is great detail in a ‘bug ’s life’. Everything is covered. The film {\color{blue}looks}({\color{red}\{glances, peeks, reviews, stares, views, expects, casts, gazes, inspects, leers, notices, observes, regards, sight, watches, sees, glimpses, reads, cares, notes\}})
great and the animation is sometimes jaw dropping. The film isn’t too terribly original.
\end{itemize}




Moreover, human evaluation is usually needed in NLP tasks, so we provide some adversarial texts generated by PCAA and show in Tables \ref{table: IMDB LSTM} and \ref{table: IMDB CNN}. In detail, we run \method attack on IMDB dataset over two victim models LSTM and word-CNN. The candidate sets are pre-specified synonym sets. Similariy, adversarial words are marked in red while original words are in blue. It is obvious that these replacements do not hurt the semantic meaning but can fool the classifiers.

\begin{longtable}{|c|c|p{9cm}|}
\caption{IMDB Adversarial Examples from PCAA on LSTM}
\label{table: IMDB LSTM}\\
\hline \hline
Class & Perturbed Class & Perturbed Texts\\
\hline
\hline
Negative & Positive & I watched this film for 45 minutes and counted 9 mullets. That's a mullet every 5 minutes. Seriously, though this film is \textcolor{red}{residing evidence}(\textcolor{blue}{living proof}) that formula works if it ain't broke,  it don't need fit in a streetwise yet vulnerable heroine, a hardened ex-cop martial arts master with a heart of gold and a serial killer with 'issues' pure magic.\\
\hline
Negative & Positive & Claustrophobic camera angles that do not \textcolor{red}{aid}(\textcolor{blue}{help}) the movie.  Too long face only shots, where you most of the time get the \textcolor{red}{hunch}(\textcolor{blue}{feeling}) that the lower half of the film is missing that the screen is cut off because there seems to be important actions going on, but you can not see them. There is anyway already too much confusion in the movie, so these viewing angles make it worse and do not contribute to artful visuals. I like artfully made movies and unconventional camera work. I can handle deep and slow movies but this one is trying too hard to be something artful and fails, in my opinion, painfully. Nothing to get attached to any of the characters because they are not worked out well enough to work out characters. More is needed than just minute long face shots. At least with this set of script director actors, I wonder whether some of the not so \textcolor{red}{decent}(\textcolor{blue}{good}) acting is due to the script and director or due to the actors. I will stay away from films both written and directed by le you for sure in the future. What an annoying film even for \textcolor{red}{person}(\textcolor{blue}{someone}) who would be interested in that part of history and for someone who spent time in Shanghai. \\
\hline
Positive & Negative & I really liked this version of 'vanishing point' as opposed to the 1971 version. I \textcolor{red}{finds}(\textcolor{blue}{found}) the 1971 version quite boring if I can get up in the middle of a movie a few times as I did with the 1971 version, then to me it is not all that great. Of course, this could be due to the fact that I was only nine at the time the 1971 version was brought out. However, I have \textcolor{red}{noticed}(\textcolor{blue}{seen}) many remakes \textcolor{red}{everytime}(\textcolor{blue}{where}) I have liked the original and older one better. I found that the plot of the 1997 version was more understandable and had basically kept true to the original without undermining the meaning of the 1971 version. In my opinion I felt the 1997 version had more excitement and wasn't so blasé boring.\\
\hline
Positive & Negative & The cast is \textcolor{red}{marvellous}(\textcolor{blue}{excellent}), the acting good, the plot interesting, the evolvement full of suspense, but it is hard to cram all those elements into a film that is barely 80 minutes long. If more time was taken to develop the plot and subplots, it would have a much better effect. Another 30 minutes of substance would have made this a very \textcolor{red}{right}(\textcolor{blue}{good}) film rather than just a good one. \\
\hline
Positive & Negative & There is great detail in a ‘bug 's life’. Everything is covered. The film \textcolor{red}{expects}(\textcolor{blue}{looks}) great and the animation is sometimes jaw dropping. The film isn't too terribly original. It 's basically a modern take on kurosawa 's seven samurai only with bugs, I enjoyed the character interaction however, and the \textcolor{red}{naughty boys}(\textcolor{blue}{bad guys}) in this film actually seemed bad. It seems that Disney usually makes their bad guys carbon copy cut outs, the grasshoppers are menacing and hopper the lead bad guy was a brilliant creation. Check this one out. \\

\hline
\hline
\end{longtable}

\begin{longtable}{ |c |c |p{9cm}| }
\caption{IMDB Adversarial Examples from PCAA on word-CNN}
\label{table: IMDB CNN}\\
\hline \hline
Class & Perturbed Class & Perturbed Texts\\
\hline
\hline
Positive & Negative & I am a college student studying A levels and need help and comments from anyone who has any views at all about the theme of mothers in film. In The Mother, whether you have gone through something similar or just want to comment and help me research more about this film, any comment would much greatly appreciated. The comments will be used \textcolor{red}{alone}(\textcolor{blue}{solely}) for exam purposes and will be included in my written exam. So if you have any views at all I’m \textcolor{red}{convinced}(\textcolor{blue}{sure}) I can put them to use and you could help me get an A. I am also studying  about a boy and tadpole. So if you have seen these films as well, I would appreciate it if you could leave comments on here on that page. Thank you. \\
\hline
Negative & Positive & This movie is so \textcolor{red}{horrendous}(\textcolor{blue}{awful}). It is hard to find the right words to describe it. At first the story is so ridiculous. A narrow minded human can write a better plot. The actors are boring and untalented. Perhaps they were compelled to play in this \textcolor{red}{dorky}(\textcolor{blue}{cheesy}) film. The camera receptions of the national forest are the only good in this whole movie. I should feel ashame because I paid for this lousy picture. Hopefully nobody makes a sequel or make a similar film with such a worse storyline.\\
\hline
Positive & Negative & This movie is wonderful, the writing, directing, acting, all are \textcolor{red}{marvelous}(\textcolor{blue}{fantastic}). Very witty and clever script quality performances by actors. Ally Sheedy is strong and dynamic and delightfully quirky really original and heart warmingly unpredicatable. The scenes are alive with fresh energy and really talented \textcolor{red}{generating}(\textcolor{blue}{production})\\
\hline
Positive & Negative & This may not be war peace but the two academy noms wouldn't have been forthcoming. If it weren't for the genius of James Wong Howe, this is one of the few films I've fallen in love with as a \textcolor{red}{infant}(\textcolor{blue}{child}) and gone back to without dissatisfaction. Whether you have any interest in what it offers fictively or not, BBC is a visual feast. I'm not saying it's his best work. I'm no expert there for sure but the look of this movie is \textcolor{red}{astounding}(\textcolor{blue}{amazing}). I love everything about it, Elsa Lanchester, the cat, the crazy hoodoo, the retro downtown Ness, but the way it was put on film is breathtaking. I even like the inconsistencies pointed out on this page \textcolor{red}{aforementioned}(\textcolor{blue}{above}) and the special effects that seem backward. Now it all creates a really consistent world.\\
\hline
Positive & Negative & Bette Midler is again divine raunchily \textcolor{red}{hilarious}(\textcolor{blue}{humorous}) in love with burlesque, capable of bringing you down to tears either with old jokes, with new dresses or merely with old songs, with more power punch than ever. All in all, \textcolor{red}{sung}(\textcolor{blue}{singing}) new ballads power, singing the good old perennial ones such as the rose ‘stay with me’ and yes even ‘wind beneath my wings’. The best way to appreciate the Divine Miss M has always been libe since this is the next best thing to it. I strongly recommended to all with a mixture of adult \textcolor{red}{extensive}(\textcolor{blue}{wide}) eyed enchantment and appreciation and a child 's mischievous wish for pushing all boundaries.\\
\hline
\hline
\end{longtable}
\end{document}